\documentclass[sigconf]{acmart}
\usepackage{caption}
\usepackage{subcaption}
\usepackage{graphicx}
\usepackage{hyperref} 
\usepackage{latexsym}
\usepackage{url}

\title{Interactive Re-Fitting as a Technique for Improving Word Embeddings}

\author{James Powell}
\email{jepowell@lanl.gov}
\orcid{0000-0002-3517-7485}
\affiliation{%
  \institution{Los Alamos National Laboratory}
  \streetaddress{P.O. Box 1663}
  \city{Los Alamos}
  \state{New Mexico}
  \country{USA}
  \postcode{87545}
}
\author{Kari Sentz}
\email{ksentz@lanl.gov}
\affiliation{%
  \institution{Los Alamos National Laboratory}
  \streetaddress{P.O. Box 1663}
  \city{Los Alamos}
  \state{New Mexico}
  \country{USA}
  \postcode{87545}
}

\date{December 9, 2019}

\begin{document}
\begin{abstract}

 Word embeddings 
 are a fixed, distributional representation of the context of words 
 in a corpus learned from word co-occurrences. 
 While word embeddings have proven to have many practical uses in natural language processing tasks, they reflect the attributes of the corpus upon which they are trained. Recent work has demonstrated that post-processing of word embeddings to apply information found in 
 lexical dictionaries can improve their quality.
We build on this post-processing technique by making it interactive. 
Our approach makes it possible for humans to adjust portions of a word embedding space by moving sets of words closer to one another. One motivating use case for this capability is to enable 
users to identify and reduce the presence of bias in word embeddings.
Our approach allows users to trigger selective post-processing as they interact with and assess potential bias in word embeddings.

\end{abstract}
\maketitle

\section{Introduction}
 Word embeddings are a contemporary manifestation of the maxim first posited by John Rupert Firth in 1957 : ``you shall know a word by the company it keeps‚'' \cite{ref:Firth1957} By capturing a target word and the distribution of words that surround it, word embeddings 
 represent a learned contextual representation for words in a corpus. 
Per-word vector representations are learned using a single layer neural network whose input is 
 an unlabeled text corpus. 
 
While word embeddings have been the leading distributional technologies, there are well known drawbacks, either as a consequence of black-box neural networks or the complexities of natural language. This has inspired a number of related techniques for improving embedding models such as in \cite{ref:FaruquiEtAl2015, ref:MrksicEtAl2016,ref:LengerichEtAl2018, ref:KuznetsovEtAl2018} many of which aim to adjust word embedding vectors using formalized human-curated semantic information.
We build on this work and introduce the idea of {\it{interactive re-fitting}} to incorporate human judgement into adjusting word embeddings in an interactive fashion.

\section{Background}
The breadth of data used to train the neural network affects the scope of the resulting word embedding. Many natural language processing machine learning applications use word embeddings constructed from a large corpus, often referred to as {\it{global word embeddings}}. For text that relies heavily on a specialized vocabulary, locally trained word embeddings may yield better results for some applications.

Word embeddings overall have two types of usage \cite{ref:WhitakerEtAl2019}: {\it{intrinsic}} tasks make use of the characteristics of the word embedding vector space to solve problems directly. For example, pair wise comparison using a distance metric can be used to find words that are semantically or syntactically relate.
{\it{Extrinsic}} tasks utilize word embeddings as features together with other machine learning algorithms. One example is the systematic selection and substitution of word embedding vectors in text sequences. The resulting multi-dimensional matrix of inputs provides richer information to neural networks performing natural language processing tasks such as text classification. This is also referred to as transfer learning and has lead to many rapid advancements in the field.

\section{Limitations of Word Embeddings}
Transfer learning has clearly lead to advancements in machine learning, 
but when neural networks are involved, it is often the case that an uninterpretable model is used to generate yet another uninterpretable model. 
Machine learning algorithms produce models that, 
try to find ``the optimal tradeoff between compression of X and prediction of Y''. This leads to the question: ``what did the model forget?'' \cite{ref:Schwartz-zivEtAl2017}  

Although it is commonly believed that most machine learning algorithms 
find a way to 
leverage the information bottleneck to eliminate noise while preserving important details, this is controverted by the findings in Mu et al. \cite{ref:MuEtAl2017} where the authors use an “all-but-the-top” post-processing technique that eliminates the top one or two principal components from the embedding space and improves the quality of the resulting embedding vectors. This highlights the broader problem of neural network uninterpretability  and inspires specific questions into how misrepresentations, spurious relationships, and biases can be incorporated into the representation and conveyed from word embeddings to other tasks.

There are some well-known shortcomings associated with word embeddings. One limitation is that word embeddings do not reliably represent antonyms and synonyms as distinct, orthogonal entities. Because of their similar usage contexts in the training corpus, antonyms may often have similar embedding vectors.\cite{ref:WhitakerEtAl2019} 

Another problem is that word embeddings will faithfully model human bias if it is found in the training corpus.
Brunet et al. \cite{ref:BrunetEtAl2019} noted that ``popular word embedding methods... acquire stereotypical human biases from the text data they are trained on.'' Learned biases need not be limited to social biases, it may also manifest as bias toward favoring a particular theory or hypothesis as though the evidence for it was stronger than it actually is, or that terms close in proximity represent evidence. A few solutions have been proposed for addressing bias, but they involve making adjustments to the input training corpus, which becomes impractical if there are many diverse biases represented. 
Post-processing of word embeddings is another way of adjusting word embeddings when the learned models fail to adequately represent word uses and meanings.  Retrofitting, counter-fitting, and other approaches are post-processing techniques that apply knowledge from external sources in order to improve the quality of word embeddings.

\section{Improving Word Embeddings}
Faruqui et al \cite{ref:FaruquiEtAl2015} consider the problem of injecting semantics into a previously trained word embedding space.  They were able to demonstrate that this improves representational semantics of the embedding vectors. Light-weight post processing of word embeddings to better represent basic relationships that can be inferred from an ontology or lexical dictionary have proven to improve embedding quality while they are also fast and efficient. The retrofitting approach can utilize synonym information to move similar terms closer to one another. The resulting vectors are adjusted so that the Euclidean distance of the vector representing the lexical entry so that it is moved closer to the terms associated with that entry in the lexical dictionary, such that the vector representing lexical entry q\textsubscript{i} moves closer to neighbor q\textsubscript{j}:

\begin{equation}
q_i=\frac{\sum_{j:(i,j)\in{E}}{\beta_{ij}}{q_j}+\alpha_{i}{\hat{q}_i}}{\sum_{j:(i,j)\in{E}}\beta_{ij}+\alpha_i}
\end{equation}

A related approach, counter-fitting, applies a reverse technique to antonyms, so that they are moved farther from opposites. \cite{ref:MrksicEtAl2016} A synonym attract function moves words with similar meanings closer to one another, while an antonym repel function moves words with opposite, or orthogonal meanings farther apart. Counter-fitting addresses the problem observed with word embeddings where semantic similarity is conflated with conceptual association. 

Inspired by this work, we  explore 
substituting
curated lexical source with real-time human judgement. Our project enables users 
to trigger two variations of the retro-fitting technique to words they want to adjust in the word embedding space. In the long term, our goal is to expose multiple post-processing techniques as  interactive tasks. We refer to these capabilities collectively as {\it{re-fitting}}.
\section{Interactive Re-Fitting}
Human interaction with word embeddings allows for the injection of human judgement into the model. Humans are better at detecting bias and can use their expert knowledge to improve or correct other relationships in the embedding space. Users can specify or emphasize particular word groupings, flag certain words as orthogonal to other words, reduce weight associated with syntactic relationships, and reduce the effects of bias. Interaction can occur after embeddings are generated via an unsupervised method. Interactions and annotations can directly affect the embedding space, but they can also be recorded as distinct activities which can later be analyzed for consistency, to improve the process, and to look for the possibility of human error. 


We define interaction as any user action that specifies a change to a relationship between any two sets of word vectors in the embedding space. These interactions are modeled as a word combination task where a user selects one or more words and uses a User Interface (UI) affordance to indicate association refinement type. Users can make a variety of adjustments using this simple approach, including spatial changes among words to reduce bias.
This would allow for iterative, crowd-sourced refinement of word embeddings, which we refer  to as interactive re-fitting of word sets. In this context, we use the word re-fitting to refer to any user-specified adjustments performed using a variety of objective functions that changes the distance among a set of embedding terms.

For this investigation, we adapted the retrofitting algorithm to function as a Web service. 
We also implemented a second, more aggressive version of retrofitting which we call {\it{retrofitting words sets}} to word embeddings. Since it is possible for the user to indicate the kind of relationship that ought to be emphasized, retrofitting to word sets employs a modification of the objective 
so that all the identified words are moved closer to one another relative to their initial positions in the embedding space. Searches into the embedding space were performed against locally trained word embeddings based on a corpus of Los Alamos National Laboratory Unlimited Released Technical Reports which includes research papers, presentations, and various other types of reports covering a period that spans mid 2012 through early 2019. Using the python NLP gensim library, the local word2vec embeddings are loaded as keyed vectors. We provided users with a Web interface allowing them to search our local word embeddings. These searches are converted to cosine similarity queries. Using the cosine metric, the search engine identifies and returns words that are most similar to the user's query. Normally this would be incorporated into another information retrieval task, but for this experiment, we simply enable users to  apply interactive refitting to words found in these results sets.
\section{Test and Results}
To demonstrate the potential of interactive re-fitting, we identified a few examples that illustrate how implicit bias might be adjusted for using this method.  Implicit bias is defined as ``implicit biases are whatever unconscious processes influence our perceptions, judgements and actions.'' \cite{ref:HolroydEtAl} Using the search tool described above, we identified terms representative of two types of implicit bias. The first set are terms often associated with gender bias, and the second reflects a historically perception that some sciences are not considered {\it{hard sciences}}. In the first case, we examined embedding vectors for the pronoun ``she'' as well as gender specific nouns ``female''‚ ``woman''‚ and ``women'' in relation to  the word ``science'' from our local word embeddings. {\it{(The authors would like to note that in fact, this form of bias was actually not reflected in our test corpus.)}} Then we simulated a human-in-the-loop standard retrofitting of target pronoun ``she'' to this set of terms to move ``she'' closer to the other words. We also used this same set of terms to perform set retrofitting of the group, in order to move all words in the set closer to one another. The original relationships, as well as standard and set retrofitting are illustrated in Figure 1. 

\begin{figure}
    \centering
    \includegraphics[width=3.25in]{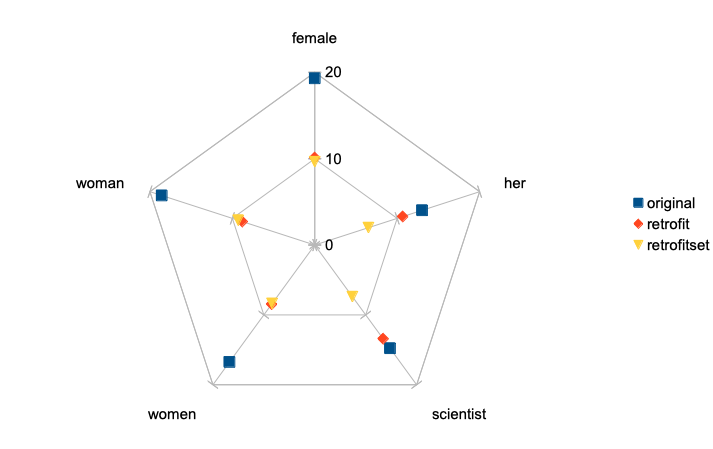}
    \caption{Original, retrofitted, and retrofitted set embedding vectors}
    \label{fig:my_label}
\end{figure}

The second example examines the relationships among hard and soft sciences in the same local embedding space. Along with the embedding vector for the general term ``science'', we evaluated labels for various  hard sciences including ``physics'', ``computer science'', ``chemistry'', and ``biology'' and terms for field sometimes perceived as soft science including ``sociology'' and ``psychology''. Indeed the soft science terms were farthest away from all of the other terms. 
Figure 2 illustrates the original positions of the terms, as well as the simulated human-in-the-loop re-fitting where both standard retrofitting and set retrofitting were applied.

\begin{figure}
    \centering
    \includegraphics[width=3.25in]{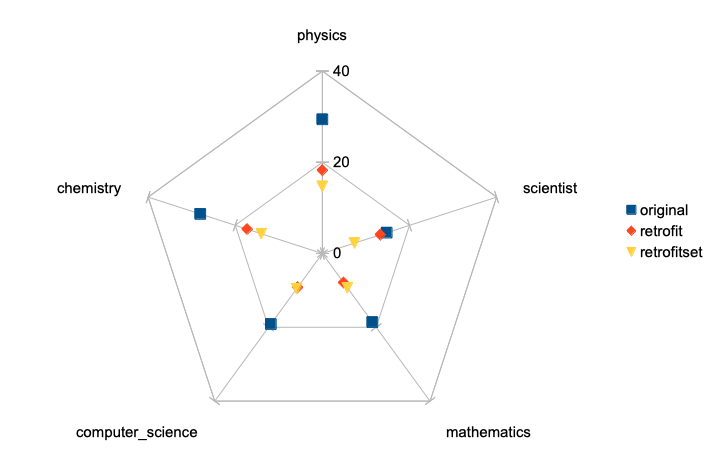}
    \caption{Changes in Euclidean distance among hard and soft sciences}
    \label{fig:my_label2}
\end{figure}

Other approaches to adjusting for bias in textual content in NLP tasks have involved extensive analysis of and modifications to the corpus. However targeting the embeddings rather than the corpus may be a more economical and re-usable alternative. Studies have shown that extrinsic tasks where word embeddings are used as input features for machine learning tasks tend to rely more on local similarities in the embedding space. Because of this, we believe that interactive identification and adjustment for these biases would have a notable impact on extrinsic tasks. This might make it possible to reduce bias by incorporating human in the loop adjustments to the embedding space. without making any changes to the corpus itself will reduce the affects of bias in a corpus, simply by performing such targeted refinements to the embedding space.

\section{Conclusions}
Post-processing of word embeddings is a proven technique for improve representational quality. It has been used to adjust relationships among words so that they better represent meanings and usages. To our knowledge, no one has attempted to apply post-processing to word embeddings in an interactive fashion. We are also not aware of any efforts to apply post-processing to word embeddings in order to correct for bias. 

Bias is inherently difficult to correct for, takes many forms, and we believe that effectively neutralizing bias requires human judgement. We have built a prototype system that enables human-in-the-loop re-fitting of word embeddings, allowing users to adjust relationships among words. We envision that this technique could be used to make precise adjustments to word embeddings in order to reduce or eliminate bias that might exist in a training corpus. We believe that these improvements would be propagated to downstream applications, thus reducing the effects of bias in a variety of machine learning NLP workflows. Future work will include exposing counter-fitting as an interactive service and evaluating the benefits of interactive re-fitting of word embeddings to extrinsic tasks.

We believe that interactive re-fitting has broader applications for improving the quality of word embeddings. Word embeddings can be used in many information retrieval tasks. Large scale deployment of re-ftting in the context of an information retrieval system represents a compelling opportunity to engage large numbers of users and directly incorporate human judgement and knowledge into word embeddings. 


\section*{Acknowledgments}
This work was funded in part by the Los Alamos National Laboratory, Laboratory Directed Research and Development and benefited greatly from insight from our colleagues, Alexei Skurikhin and Reid Porter, and the resources provided by the Research Library at Los Alamos National Laboratory.

\bibliography{acl2018}
\bibliographystyle{acl_natbib}

\end{document}